%% file: mwe.tex
\RequirePackage{fix-cm}

\documentclass[smallextended]{svjour3}
\smartqed  
\usepackage{graphicx}
\usepackage{adjustbox}
\usepackage{subfig}
\usepackage{multirow}
\usepackage[table,xcdraw]{xcolor}

\pdfoutput=1

\usepackage{float}
\journalname{The Journal of Supercomputing}
\begin{document}

\title{A framework for robotic arm pose estimation and movement prediction based on deep and extreme learning models}


\author{Iago Richard Rodrigues \and
        Marrone Dantas \and
        Assis Oliveira Filho \and
        Gibson Barbosa \and
        Daniel Bezerra \and
        Ricardo Souza \and
        Maria Valéria Marquezini \and
        Patricia Takako Endo \and
        Judith Kelner \and
        Djamel H. Sadok 
}


\institute{F. Author \at
              first address \\
              Tel.: +123-45-678910\\
              Fax: +123-45-678910\\
              \email{fauthor@example.com}
           \and
           S. Author \at
              second address
}

\date{Received: date / Accepted: date}

\maketitle

\input{sections/abstract}

\input{sections/introduction}
\input{sections/related-works}
\input{sections/proposed-framework}
\input{sections/materials-methods}
\input{sections/results}

\input{sections/conclusion}

\section*{Declarations}
\begin{itemize}
    \item \textbf{Data availability:} The datasets generated during and/or analysed during the current study are available from the corresponding author on reasonable request.
    \item \textbf{Conflict of interest:} The authors declare that they have no conflict of interest.
\end{itemize}

\begin{acknowledgements}
This work was financed in part by the Conselho Nacional de Desenvolvimento Científico e Tecnológico (CNPq), Fundação de Amparo a Ciência e Tecnologia de Pernambuco (FACEPE), Coordenação de Aperfeiçoamento de Pessoal de Nível Superior (CAPES), and Research, Development and Innovation Center, Ericsson Telecommunications Inc., Brazil.
\end{acknowledgements}

\bibliographystyle{unsrt}

\bibliography{mwe2}   

\end{document}

%% file: sections/abstract.tex
\begin{abstract}

Human-robot collaboration has gained a notable prominence in Industry 4.0, as the use of collaborative robots increases efficiency and productivity in the automation process. However, it is necessary to consider the use of mechanisms that increase security in these environments, as the literature reports that risk situations may exist in the context of human-robot collaboration. One of the strategies that can be adopted is the visual recognition of the collaboration environment using machine learning techniques, which can automatically identify what is happening in the scene and what may happen in the future. In this work, we are proposing a new framework that is capable of detecting robotic arm keypoints commonly used in Industry 4.0. In addition to detecting, the proposed framework is able to predict the future movement of these robotic arms, thus providing relevant information that can be considered in the recognition of the human-robot collaboration scenario. The proposed framework is based on deep and extreme learning machine techniques. Results show that the proposed framework is capable of detecting and predicting with low error, contributing to the mitigation of risks in human-robot collaboration.

\keywords{Robotic arm pose estimation \and Movement prediction \and Deep learning \and Extreme Learning Machine}

\end{abstract}

%% file: sections/introduction.tex
\section{Introduction}


The usage of robots to perform human tasks has been increasingly applied in the production process, manufacturing, and other areas of the industrial sector \cite{goel2020robotics}. Robots are also assisting humans in undertaking specific tasks in what is referred to as human-robot collaboration (HRC) \cite{ajoudani2018progress}. Currently HRC spans diverse contexts, including manufacturing, homes, offices, and hospitals \cite{bauer2008human}. More significantly, HRC has been introduced in Industry 4.0 manufacturing activities. This triggered interest and set a new trend in the research community seeking to evaluate the impact and ensure efficient use of HRC \cite{Microsoft:2019trends}. Advocates of this technology claim that it benefits the production process by reducing the time to execute some complex activities traditionally allocated to humans only.

The present study looks at HRC in the context of the maintenance of computer network equipment, such as a radio base station (RBS), as the one described in \cite{reis2021gripper}. Cellular network operators allocate considerable human and financial resources for the maintenance of RBS equipment. This is a task that requires frequent visits by technicians to geographically distributed sites over small and large urban environments. An Telecom operator must ensure high RBS availability in order to avoid financial loss. The inherent cost of these visits and the time it takes to locally perform maintenance tasks are a concern. This problem is likely to worsen with the introduction of a large number of small cells as planned by the 5G standard \cite{thors2017time}. Collaborative robots (COBOTS), specifically robotic arms, can be successfully deployed jointly with technicians to automate many of the RBS related maintenance tasks. Their collaborative tasks may cover a range of activities including faulty cable verification and their replacement, insertion and removal of cables into and from network device ports, device configuration, among others.

Despite the many apparent advantages, the direct contact between humans with robots can pose some risks. One of these is collision, see \cite{vasic2013safety}. Such events may lead to accidents that can cause irreparable physical damage to humans \cite{rodrigues2021modeling}. Robots are also likely to suffer from collisions, possibly leading to financial losses to owners. There is as a result a need to design, deploy and evaluate new safety mechanisms that govern HRC and ensure its safety. Anticipating and mitigating risks in HRC certainly provides a promising solution to this problem.

Considering the previously identified research gaps, the need for safe HRC and the promising high accuracy of DL models for detection tasks, the present work develops and evaluates a new framework for the accurate detection of robot poses and the prediction of future movement. The proposed framework consists of the following two parts:

\begin{enumerate}
    \item Robotic arm pose estimation: relies on re-training a CNN model named SCNet-50-V1-d \cite{liu2020improving} for the pose detection task of the robot through regression. The SCNet-50-V1-d model uses self-calibrated convolutions (SCConvs) and enjoyed a great deal of success in a wide range of application including image classification, object detection and instance segmentation. In this work, we also demonstrate the efficacy of using SCConvs for regression tasks. In addition, we improve pose estimation through the use of the extreme learning machine (ELM) neural network \cite{huang2006extreme}, which when combined with CNN models tends to provide a better learning outcome \cite{rodrigues2021convolutional}.
    \item Robotic arm movement prediction: contemplates the training a long-short term memory (LSTM) \cite{hochreiter1997long} and gated recurrent unit (GRU) models \cite{cho-etal-2014-learning} to predict the future movement of a robotic arm.
\end{enumerate}

The framework is instantiated with several DL models that support pose detection and prediction of future robot movement. These are evaluated and compared in a well-controlled scenario, as defined by Silva et al. \cite{silva2020assessing}. Despite being a well-controlled setting, the adopted testbed includes real equipment such as the UR-5 robotic arm, networking and RBS devices installed in a rack. The scenario also inserts a person  responsible for performing collaborative activities with the robotic arm. These activities are recorded through a strategically positioned camera. We also annotate the data to carry out the regression of the robot's keypoints and obtain the current robot pose. We compared all the evaluated models using the two performance metrics: mean squared error (MSE) and mean absolute error (MAE) metrics. The obtained result demonstrate the suitability of using the proposed framework as an efficient solution for the detection and prediction of the future movement of a robotic arm while suffering a very low error.

\subsection{Organization of this article}

The remainder of this article is organized as follows. Section \ref{sec:relatedWorks} presents similar works in the robot keypoint detection field. Section \ref{sec:proposed_system} presents the proposed framework and details its two main modules, namely, robotic arm pose estimation and robotic arm movement prediction. Section \ref{sec:materials_and_methods} describes the methods adopted in this work for developing the experimental scenario, supported image acquisition process, used DL models, used parameters and metrics, and finally model validation. Section \ref{sec:results_and_discussion} explains the obtained results 0along with the experiments performed in this study. Finally, Section \ref{sec:conclusion} concludes this article and suggests some future research directions.

%% file: sections/related-works.tex
\section{Related Works}
\label{sec:relatedWorks}

Across the literature, there are several technologies that can actually contribute towards detecting and mitigating risks in HRC \cite{robla2017working}, \cite{lasota2017survey}, \cite{silva2020assessing}, \cite{zhang2020recurrent}, \cite{anvaripour2019collision}, \cite{maceira2020recurrent}. Among these, we highlight the special role of machine learning (ML), which provides intelligent systems to solve the most diverse engineering problems, including HRC. In particular, we draw the attention to an emerging sub-field of ML in the literature, deep learning (DL) \cite{lecun2015deep}. DL has proven to beat the state of the art in solving most common problems involving ML, such as the image classification challenge in the ImageNet database \cite{deng2009imagenet}, \cite{tan2019efficientnet}. Furthermore, DL has gained notable prominence in regression, clustering, among other tasks, through its use of convolutional neural networks (CNN) \cite{liu2017survey} and recurrent neural networks (RNN) \cite{lipton2015critical}.

As a result, there are several approaches in the literature that use DL models for risk mitigation in HRC \cite{silva2020assessing}, \cite{sharkawy2020human}, \cite{park2020learning}. Nonetheless, many of the already listed works are limited to carrying out the immediate detection of collision events. They do not predict eminent risk situations before they happen. To reduce damage to workers and agents involved in close collaboration scenarios with robots, one should ideally train new models that predict risk situations and analyze relevant data to predict possible risks in a timely manner and prior to taking place \cite{zhang2020recurrent}, \cite{anvaripour2019collision}, \cite{maceira2020recurrent}.

The authors are of the view that there are limited contributions that take into consideration the detection of human intention to estimate the likelihood of future risks. Such works suffer from a limited scope as they focus on detecting human movement \cite{zhang2020recurrent}, \cite{anvaripour2019collision}, \cite{maceira2020recurrent} and do not combine this with robot detection and the analysis of its movement. Note however that there are literary works that tackle robotic arms detection in an isolated context. Unlike these, this study develops and evaluates detection mechanisms for robots to assess their proximity, speed, collision risks, future movement, etc., in a HRC scenario. These assessments should provide a more robust solution for risk mitigation in HRC. 

Pose estimation (or detection) is widely used in several computer vision applications such as medical assistance, games, and human intention detection. All previously cited applications refer to human pose estimation. Human pose estimation consists of detecting the human body keypoints through regression models. In a similar view, robot pose estimation consists of detecting the robot keypoints through regression models, these keypoints are the joints that compose the robot skeleton. With the advance of deep learning studies, there has been a notable advance in robot pose detection with several CNN applications with different types of robots. Robot pose detection aims to obtain the keypoints which represent the robot joints, thus constituting its pose. Bellow, we present recent papers on robot pose detection using CNNs.

Miseikis et al. \cite{miseikis2018robot} present an approach for detecting robot poses in RGB images. Their method consists of a cascade of two different convolutional networks. The first network performs the segmentation of the robot, with an architecture reminiscent of AlexNet \cite{krizhevsky2012imagenet}. In contrast, the second network is responsible for mapping the robot's key points from the mask extracted with the first network. The UR-5 robotic arm pose detection reached an accuracy of 98.1\% and outperformed that of UR-3 and UR-10 with 93.1\% and 92.8\%, respectively. 
In a similar research, Mivseikis et al. \cite{mivseikis2018transfer} propose the use of transfer learning to detect various tasks (as \cite{miseikis2018robot}), including the detection of keypoints of another type of robot, the Kuka LBR iiwa. The approach achieved an accuacy of 97.3\% for detecting the new robot's masks. This demonstrates that transfer learning also contributes to detecting keypoints of robots, even though there are fewer examples used for training the neural network.

Zhou et al. \cite{zhou20193d} also propose an architecture for detecting the pose of robots from RGB images. The architecture is divided into two parts, Base-Net and Pose-Net. In Pose-Net, the positions of the keypoints in the image are initially estimated to guide training. Base-Net obtains the 3D position of the robot base by calculating the depth and using the robot base position in the image evaluated by proposed system. 

Lee et al. \cite{lee2020camera} present an approach to estimate the pose of the robot using a single RGB image. As with similar previous works, a convolutional network is used to detect the robot's keypoints in artificial images due to the lack of the availability of real data reported by the authors. However, unlike previous works, the projection of keypoints is in 2D. Perspective-n-point (PnP) is then used to retrieve the camera extrinsic, allowing the approach to dismiss the need from an offline calibration step from a single frame.

Heindl et al. \cite{heindl2019learning} use a recurrent convolutional network based on ConvGRU-type convolutions \cite{ballas2015delving}. Despite the promising approach and reaching a mean average precision or mAP (threshold 50\%) of 88.7\%, all images used in their entirety were artificially generated to train the proposed architecture. Although the paper presents inferences in authentic images, there are no experiments or other evidence that the proposed approach can work in real scenarios. The idea of using artificial images was also adopted in previous work \cite{heindl20193d} by the same authors in \cite{heindl2019learning}, however, the neural networks were modeled with vanilla convolutions this time.

With regard to robot pose detection, there is a need to consider the importance of robots' keypoints in the context of safety for HRC. Note also that none of the reviewed works considers the analysis of robots' keypoints information into a recurrent neural network in order to classify or predict robots' movements. In a typical HRC scenario, it is necessary to understand and classify/predict the robot's intention. This approach should help to avoid or decrease the impacts of collision/accident events.

Considering the current state-of-art, we can cite the following main contributions of our work:
\begin{itemize}
    \item This work develops a new framework designed to predict robotic arm pose detection and its future movement in the context of HRC using DL, ELM and one 2D camera in a well controlled scenario.
    
    \item Our framework is applied and evaluated in the proposed controlled setup that highly mimics a real scenario, generating real data during the experiments for analysis. Unlike most existing related works, this one did not used synthetic images, e.g. robot located in random scenarios. It is also important to highlight that in our experiments the camera calibration process was not necessary. The additional steps part of the calibration processr, such as checkerboard position, camera positioning, acquisition of intrinsic and extrinsic parameters are error sensitive. Our goal is to use a generic environment where we could use devices present in-loco.
    
    \item To the best of our knowledge, this work is first to use ELM to solve a regression problem using features generated by a neural network that uses self-calibrated convolutions (SCNet). It is first to obtain robot pose detection using a convolutional extreme learning machine (CELM) approach. We demonstrate through extensive experiments that the DL model SCNet50 used in conjunction with the ELM method provide better detection results in our scenario.
    
    \item Another important contribution of this work is the proposition of deep learning models to predict the future movement of the robotic arm. Models of this type had not been previously proposed in the literature to tackle this problem. There are works that apply DL only to human movement prediction models in the HRI context. Using DL, it becomes possible to analyze the future movement of the robot. Furthermore, DL models will also be considered, as part of future work, in the analysis of collision risk between humans and robots.
    
    \item Finally, the last contribution of this work is that the proposed framework is easy to deploy. The testbed  relies on infrastructure that is commonly found in workspaces including cameras, robotics arms, and networking devices.
    
\end{itemize}

%% file: sections/proposed-framework.tex
\section{Proposed framework}
\label{sec:proposed_system}

This section provides detailed information on the composition of the framework proposed in this work. Its adopted models, their comparison and experimental parameters will be presented in Section \ref{sec:materials_and_methods}. 

\begin{figure}[H]
    \centering
    \resizebox{\columnwidth}{!}{\includegraphics{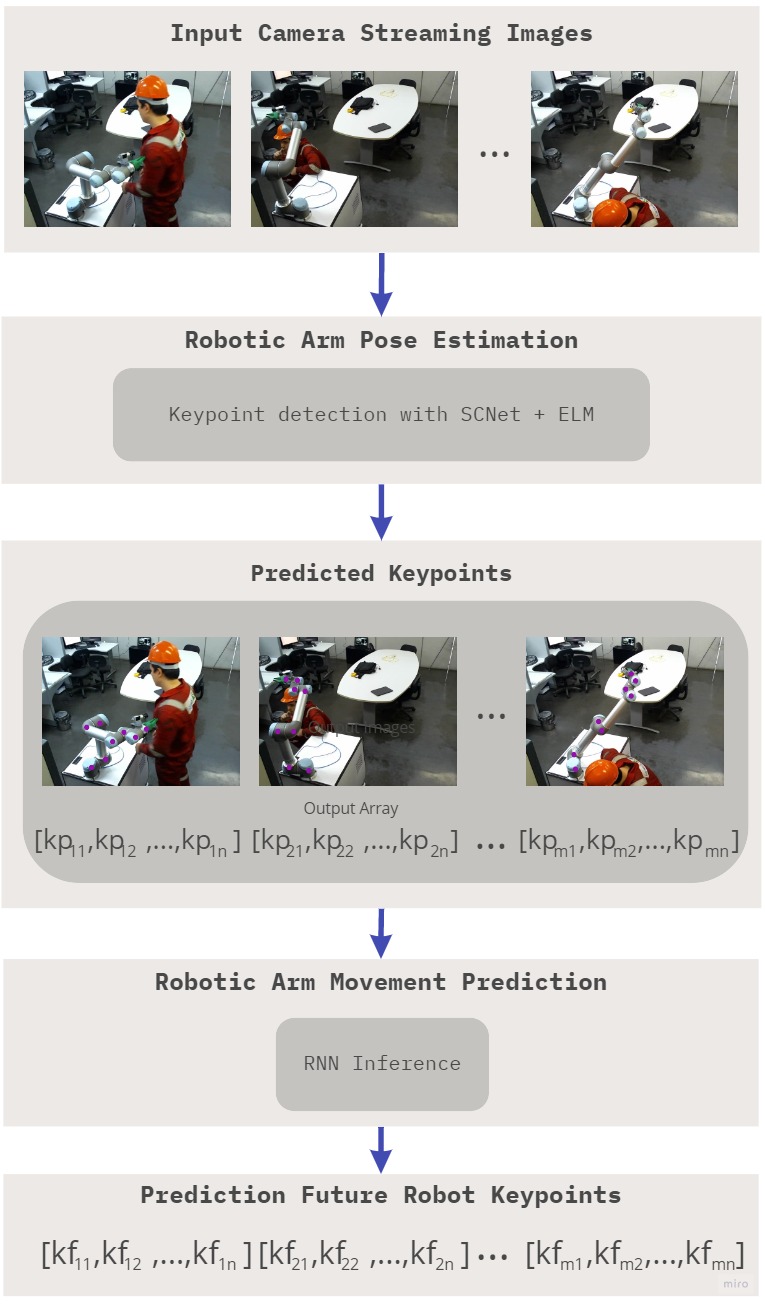}}
    \caption{Framework for robotic arm pose estimation and prediction of future movement.}
    \label{fig:proposed_general_scheme}
\end{figure}

As illustrated in Fig. \ref{fig:proposed_general_scheme}, our framework consists of two main modules. The first one is the robotic arm pose estimation module. It uses a new proposed neural network model that combines the self-calibrated CNN (SCNet-50) and ELM. The second module is the robotic arm movement prediction. It executes well known RNN models to make future inferences. The next two subsections detail the operation of these two modules.

\subsection{Robotic arm pose estimation}

\begin{figure}[ht!]
    \centering
    \resizebox{\columnwidth}{!}{\includegraphics{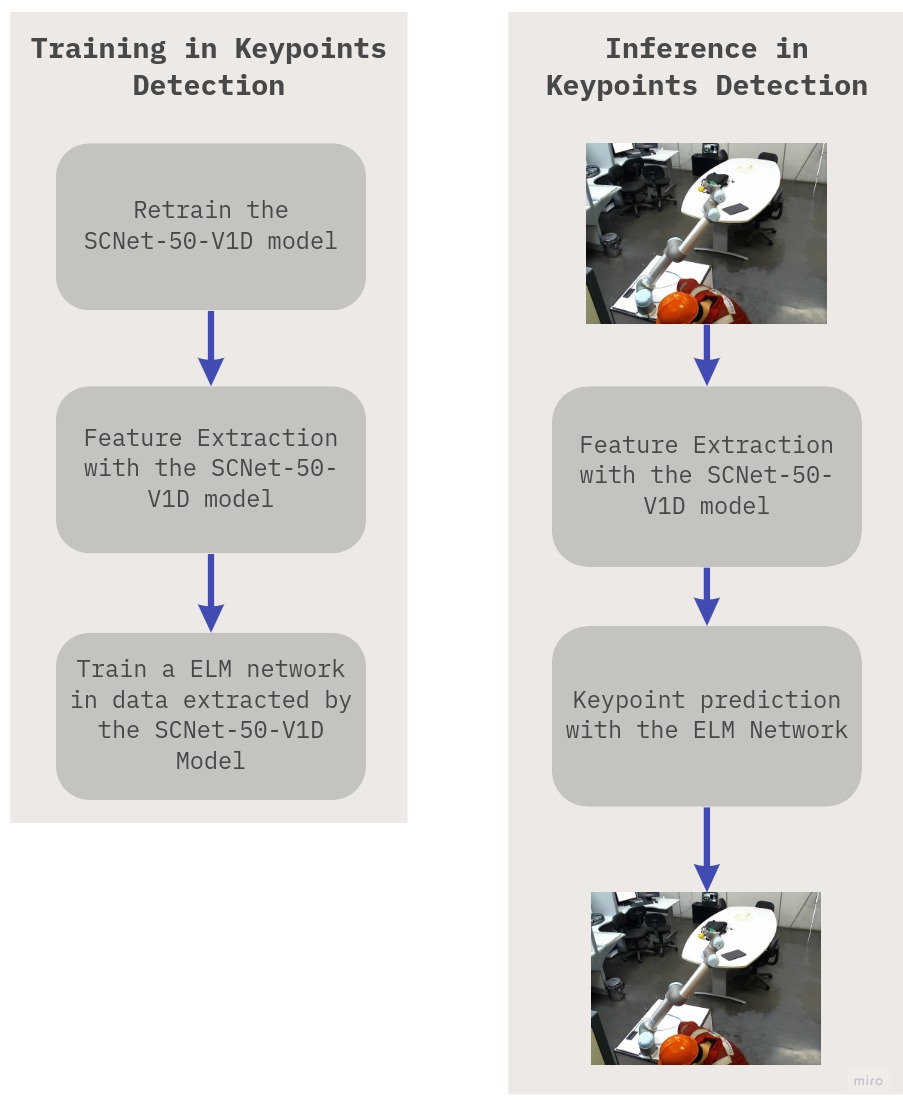}}
    \caption{Module for robotic arm pose estimation.}
    \label{fig:proposed_module1}
\end{figure}

Fig. \ref{fig:proposed_module1} depicts this module. The framework proposes a new neural network model for detecting keypoints used to estimate the pose of the robotic arm.

The first part of the proposed new model, responsible for the training in keypoints detection, uses the SCNet-50 network with self-calibrated convolutions (SCConvs). As mentioned by Liu \textit{et al.} \cite{liu2020improving}, in the SCConvs ``\textit{each spatial location is allowed to not only adaptively consider its surrounding informative context as embeddings from the latent space functioning as scalars in the responses from the original scale space, but also model inter-channel dependencies}''. This allows SCConvs to learn better discriminative representations by adding basic convolution transformations, identity, and sigmoid functions to each layer \cite{liu2020improving}.

The SCNet-50 version that we use has a modification also proposed by the SCConv authors \cite{liu2020improving}, the SCNet-50-V1D, which implements the original SCNet-50 network with modifications inspired by bag of tricks for CNNs \cite{he2019bag}. According to Liu, SCNet-50-V1D replaces the original $7\times7$ convolutions by three $3\times3$ convolutions, and in the downsampling block, a $3\times3$ average pooling with stride 2 is added before the convolution operations, whose stride is changed to 1 \cite{liu2020improving}. 

These modifications follow some of the bag of tricks paper \cite{he2019bag}. We first take the SCNet-50-V1D model trained in the ImageNet database \cite{deng2009imagenet} and then attempt to use it for transfer learning in our work. Here, we we re-train all the weights of the network in our own database (described in Section \ref{sec:dataacpro}) for robotic arm pose detection. The training process consists of removing the last fully connected layer and adding a new linear layer containing 16 activations. There are 16 activations because in this work we aim to make the regression of the eight keypoints that form the pose of the robotic arm, for each keypoint there is a coordinate $(x,y)$. 

The second part of the proposed new model, responsible for the inference in keypoints detection, uses ELM to improve the detection of poses of the robotic arm. According to a survey carried out by Rodrigues \textit{et al.} \cite{rodrigues2021convolutional}, ELM can be used together with CNNs to improve the results in classification tasks and pre-trained CNNs can be used for feature extraction. ELM models were trained with the extracted features, providing better accuracy results compared to using only CNNs.

In this work, we hypothesize that the pose estimation generated by the SCNet-50-V1D model can be improved by using the ELM network for regression. Also, we hypothesize that it is possible to have better convergence and better error results in detection for robotic arm pose estimation, considering that ELM proved itself accurate in another regression problem, according to Huang \textit{et al.} \cite{huang2015trends}.

For training, we perform the feature extraction process through our CNN network already trained in our dataset. Then we use the ELM network for a new training with the extracted features. As for inference, with all the aforementioned models, we also execute the feature extraction process using the SCNet-50-v1-d network. The already trained ELM network will also predict the keypoints, which together build the pose of the robotic arm. As a result, for each image processed, our proposed learning scheme produces outputs containing all 8 $(x,y)$ points, corresponding to the robotic arm joints. We name our proposed model as SCNet-50-V1D+ELM to make the robotic arm pose estimation.

Since we now have the model SCNet-50-V1D+ELM to make inferences about the pose of the robotic arm, there is now also a need to make inferences from future frames to predict its movement. This type of forecast can facilitate the understanding of what is happening in collaborative activities. To this end, the proposed framework is also covered with the robotic arm movement prediction module, which we will explain later.

\subsection{Robotic arm movement prediction}

\begin{figure}[ht!]
    \centering
    \resizebox{\columnwidth}{!}{\includegraphics{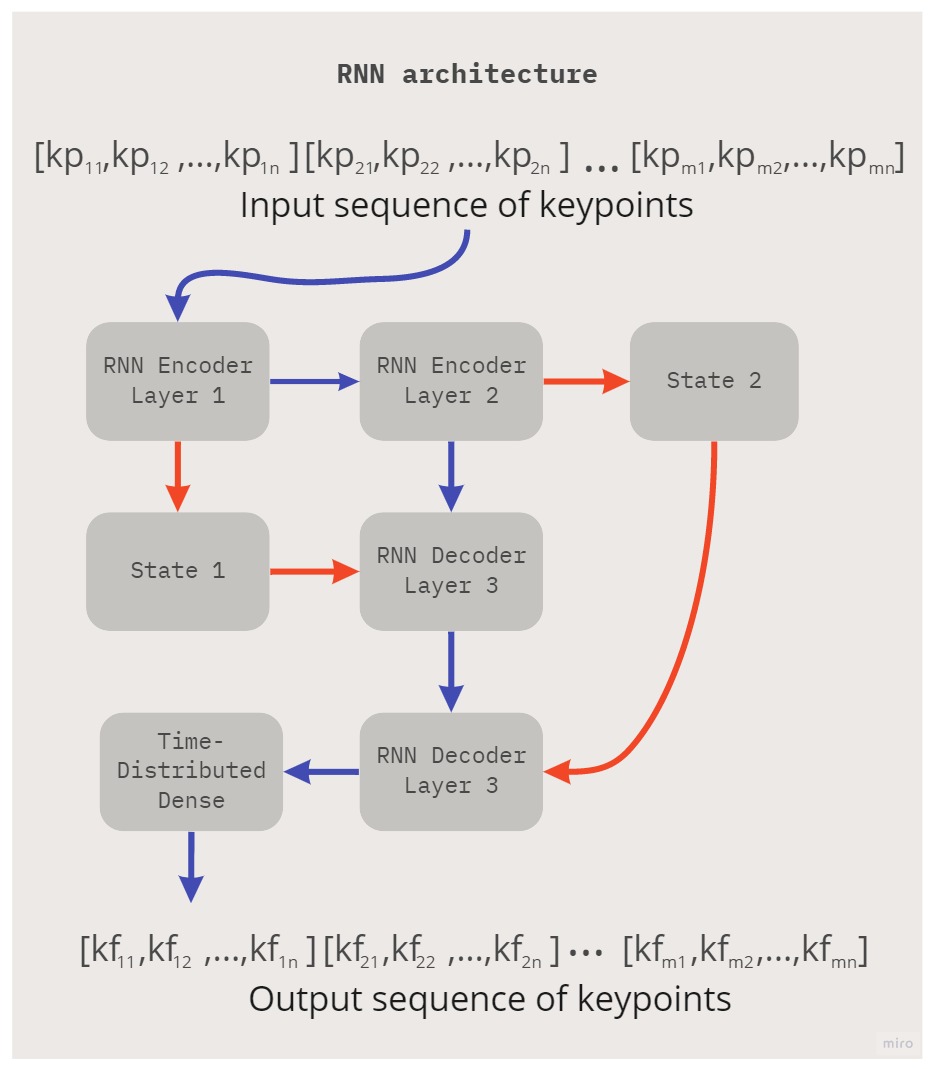}}
    \caption{Detailed view of the module for robotic arm movement prediction. Orange arrows refer to copy operations, and blue arrows refer to the normal feedforward process of the neural network.}
    \label{fig:encdecrnn}
\end{figure}

This module proposes the use of several RNN models to predict future robotic arm movements. As part of this task, we consider two RNN models: LSTM and GRU. These models are frequently used for time series prediction and. Overall, they outperform traditional ML algorithms, depending on the applications at hand \cite{ribeiro2020short}. It is important to highlight that we are proposing a double-stacked RNN (with LSTM and GRU) with encoders and decoders \cite{cho2014learning} (Fig. \ref{fig:encdecrnn}), for better learning of long term data, storing in memory various data from the past, and considering them in future prediction. There are two encoder layers, in which we store the sequences for subsequent use by the decoder layers. At last, there is a time-distributed and dense layer that makes predictions.

For each set of images $In$ supplied to the first detection module, this will return an array of dimension $(n\times16)$, where $n$ corresponds to the number of input images, and 16 corresponds to the number of coordinates generated for the identification of the robotic arm in the $In$ images. This matrix corresponds to a time series containing the keypoint coordinates of the robotic arm for the entire set of input images.

Thus, the initial predicted data is sequential (of size $(n\times16)$ containing just robotic arm poses. The RNN models are used to make future inferences with this initial data. They return another sequential data containing the predictions of the robotic arm movement. The out sequential data has a size of $(f\times16)$, where $f$ is the number of future frames set to be predicted. With these data provided, it is possible to verify where the robot is moving, enabling the creation of policies in the robot's operation. An example of an applicable policy is to stop the robot when it is moving towards a human. The output size $f$ is an adjustable parameter, together with the $n$ parameter, both are defined in Section \ref{sec:rbtmvtpred}.

%% file: sections/materials-methods.tex
\section{Materials and methods}
\label{sec:materials_and_methods}

This section provides insights on the way the experiments were setup and executed.

\subsection{Experimental scenario}
\begin{figure}[!ht]
    \centering
    \resizebox{\columnwidth}{!}{\includegraphics{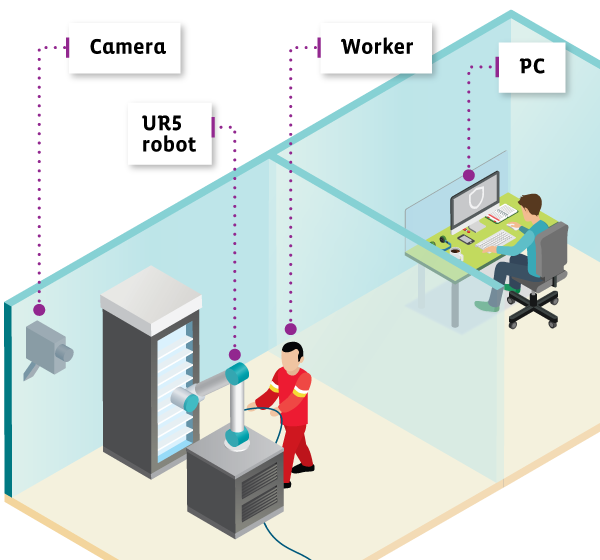}}
    \caption{System general architecture (same scenario of Silva et. al \cite{silva2020assessing}).}
    \label{fig:scenario}
\end{figure}

Fig. \ref{fig:scenario} presents our well-controlled experimental testbed implemented within a laboratory as well as all its main elements. The adopted scenario is one where a worker collaboratively interacts with a robotic arm that performs maintenance activities on an RBS network rack.

The robotic arm used in our scenario is an UR-5 \cite{ur2021robots}. It performs maintenance operations such as adding, removing, or exchanging network cables in a rack similar to the one used in an RBS site. In addition, a camera is strategically placed to capture all activities performed in our well-controlled scenario. With the interaction of the robotic arm, all images are captured at a 21 FPS rate in high definition by a webcam and processed by a nearby computer. This machine also hosts the software for training using deep learning models and making inferences. 

Table \ref{tab:deviceSpecifications} presents, in detail, the characteristics of the camera and computer used.

\begin{table}[!ht]
\centering
\caption{Specifications of the camera and computer used in this work.}
\label{tab:deviceSpecifications}
\resizebox{\columnwidth}{!}{
\begin{tabular}{lll}
\hline
\textbf{Device} & \textbf{Specification} & \textbf{Value} \\ \hline
Camera& Type            & Webcam \\
& Model               & Logitech C270 \\
& Transmission rate   & 21 fps \\
& Image Quality       & HD - 1080p \\ \hline
Computer& OS          & Ubuntu 18.04                         \\
& Processor   & Core i7-2600 CPU - 3.4 GHz           \\
& RAM         & 16GB                                  \\
& System type & 64 bits                              \\
& Video card  & GeForce GTX 1060 6GB                 \\ \hline
\end{tabular}}
\end{table} 

\subsection{Data acquisition and processing}
\label{sec:dataacpro}

All scene images captured by a single camera are next processed by a dedicated computer. In order to ensure the capture of representative scenes with HRC risk, we established some activity protocols to be followed by the robot and humans to simulate collaborative activities. Based on our previous work \cite{silva2020assessing}, we defined the following activities:

\begin{itemize}
     \item a worker making movements inside and outside the robot's workspace;
     \item a worker performing delivering or exchanging cables to the robot;
     \item accidental collision and contact situations caused by the worker or the robot by teleoperation;
\end{itemize}

The captured the images are stored in video format (.AVI) during the recording process. At the end of the recording, we extracted all frames from the video which resulted in a total of 23,135 frames (approximately 18 minutes with 20 FPS). We use the OpenCV library in Python to perform these two procedures.

\subsubsection{Data for robot pose estimation}

With the obtained frames at hand, we next selected a small number of frames to annotate robot keypoints. We considered 1,000 frames representing around one frame per second. We decided to select frames in this way to avoid excessive redundant information that does not contribute to the efficiency of the learning process. It is known that frames are given with temporal interdependence, and they generally are similar to each other. As a result, it is often not necessary to use all frames for the detection task.

\begin{figure}[ht!]
\centering
\begin{tabular}{@{}ccc@{}}
\subfloat[]{\includegraphics[scale=0.57]{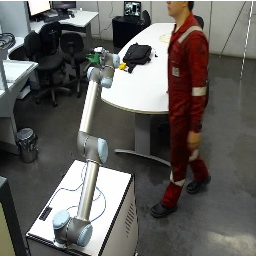}} &
\subfloat[]{\includegraphics[scale=0.57]{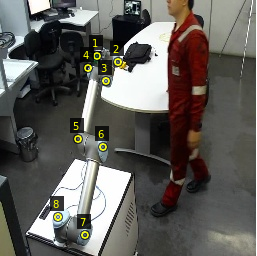}}
\end{tabular}
\caption[]{Example of robot keypoints annotation with the VGG Image Annotator software.}
\label{fig:via}
\end{figure}

To detect the pose of the robotic arm, it is necessary to define, through annotations, the keypoints that compose the robotic arm. To annotate the data for robotics arm pose detection, we use the VGG Image Annotator \cite{dutta2019via}. Fig. \ref{fig:via} shows an example of how the data was annotated.

Using the VGG Image Annotator (VIA), we annotate eight points that correspond to the robot's joints in all frames. The generated output by the tool is in JSON format. After the frame selection process, we obtained a number of 508 annotated images with the pose of the robotic arm.

\subsubsection{Data for robot movement prediction}

With regard to the data used for training the recurrent models, we did not perform frame selection because for movement prediction one needs to maintain a temporal dependency in the data. But, one problem emerges as a result of this decision, there are more than 20,000 frames that make the ground truth and which can all be used to train the recurrent models. 

\begin{figure}[ht!]
    \centering
    \resizebox{\columnwidth}{!}{\includegraphics{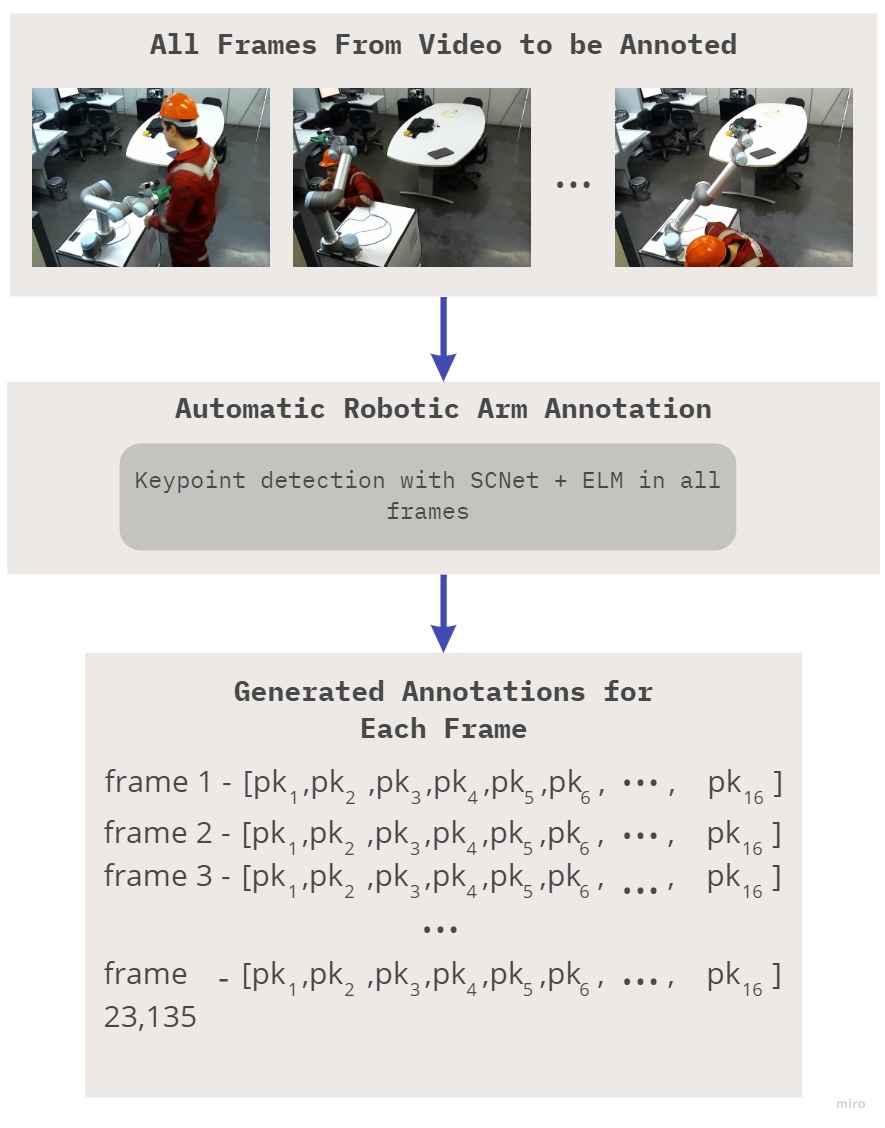}}
    \caption{Automatic process for keypoint annotation proposed in this work.}
    \label{fig:datamovementpred}
\end{figure}

We hypothesize that our proposed model for robotic arm pose estimation (SCNet-50-V1D+ELM) can extract the keypoints with an acceptable level of precision, so we assume that it is the source of our annotations. We make predictions with the SCNet-50-V1D+ELM architecture in all frames extracted, then create a dataset for robotic arm movement prediction. In other words, we obtain the position of the robotic arm in all frames of the video. With that, we created a new time series dataset containing the eight robotic arm points, with coordinates $(x,y)$. The result totals to 16 coordinate information representing the robotic arm, for each frame, as shown in Fig. \ref{fig:datamovementpred}. After runnning the automatic annotation process using the SCNet-50-V1D+ELM model, we obtained 23,135 sequential images annotated with the pose of the robotic arm.

\subsection{Experimental settings}

\subsubsection{Robot pose estimation}
Section \ref{sec:proposed_system} describes the main model for detecting the pose of the robotic arm, namely, the SCNet-50-V1D+ELM model . To prove the model's effectiveness, we compare it with some state-of-the-art DL models, such as AlexNet \cite{krizhevsky2012imagenet}, SqueezeNet \cite{iandola2016squeezenet}, VGG-11 \cite{SimonyanZ14a}, ResNet-34 \cite{he2016deep}, and DenseNet-121 \cite{huang2017densely}. For all of these models, we removed the last fully connected (or dense) layer and replaced it with another dense layer with linear activation, now containing 16 neurons responsible for predicting the keypoints of the robotic arm. During the training process, we retrained all models using their pre-trained weights obtained from ImageNet.

We trained all the models mentioned above (in addition to our proposed model) in 500 epochs with the Adam optimizer. A learning rate of 0.0001 and batch size of 8 were used. Data validation for this training process was carried out using cross-validation with 5 folds. Considering that we have a regression problem, we used the loss function of mean squared error (MSE) to calculate the error in the propagation of neural networks. 

For the ELM network, we have four different kernel activation functions, including the radial basis function (RBF) \cite{huang2005extreme}, RBF with L2-norm (RBF-L2) \cite{huang2005extreme}, hyperbolic tangent (Tanh) \cite{baraha2017implementation}, and the linear function. For all ELM evaluations, we varied the number of neurons between 100 and 1000 with a step of 50.

In addition to its role as a loss function, the MSE was also used as an evaluation metric. In addition, the proposed framework adopted the mean absolute error (MAE) as a secondary metric to evaluate the models. Eqs. \ref{eq:mse} and \ref{eq:mae} detail the definitions of the adopted MSE and MAE metrics.

\begin{equation}
    \label{eq:mse}
    MSE = \frac{1}{N} \sum_{i=1}^{N} (y_{i} - p_{i})^{2}
\end{equation}

\begin{equation}
    \label{eq:mae}
    MAE = \frac{1}{N} \sum_{i=1}^{N} |y_{i} - p_{i}|
\end{equation}

Where $N$ is the number of samples in the set, $y_{i}$ corresponds to the current real value, and $p_{i}$ is the predicted value by the regression model. 

\subsubsection{Robot movement prediction}
\label{sec:rbtmvtpred}

We train the LSTM and GRU models with the Adam optimizer over 500 epochs. The learning rate equals 0.0001, and the batch size is 4096. Another variation in the experiments with the recurrent models is the window size of the data used to make inferences (past window - $n$) and the window size of the predicted data (future window - $f$). We designed grid search experiments to analyze the impact of past and future window variation in the prediction results. Table \ref{tab:gridsearchExp} shows the parameters for the grid search.

\begin{table}[!h]
\centering
\caption{Grid search parameters used in the robotic movement prediction.}
\label{tab:gridsearchExp}
\begin{tabular}{ll}
\hline
Parameter     & Values                    \\ \hline
Past window ($n$)   & 10, 20, 30, 45, 60        \\
Future window ($f$) & 1, 5, 15, 30, 60, 90, 120 \\ \hline
\end{tabular}
\end{table}

%% file: sections/results.tex
\section{Results and discussion}
\label{sec:results_and_discussion}

This section presents the main results for both pose detection and movement prediction of a robotic arm.

\subsection{Robotic arm pose estimation}

First, we present the experimental results for the robot pose detection. Table \ref{tab:resultsA} shows the results obtained for robot pose detection using DL models, considering the mean and standard deviation of MSE and MAE reported for each DL model in the cross-validation training. All MSE and MAE results reported in this work are real values (difference in pixels). That is, they are not normalized.

\begin{table}[ht!]
\centering
\caption{MSE and MAE results reported by each DL models for robotic arm pose detection.}
\label{tab:resultsA}
\begin{tabular}{lll}
\hline
Model        & MSE         & MAE                                    \\ \hline
AlexNet      & 13.65±8.27  & 2.37±0.29         \\
DenseNet-121 & 07.33±4.67  & 1.67±0.22 \\
ResNet-34    & 12.27±7.61  & 2.31±0.91          \\
SCNet50-V1D  & \textbf{04.12}±\textbf{3.20}  & \textbf{1.24}±\textbf{0.13} \\ 
SqueezeNet   & 08.91±4.91  & 1.90±0.17  \\
VGG-11       & 27.34±3.56  & 3.92±0.20  \\ \hline
\end{tabular}
\end{table}

By analyzing the results, we can verify that the SCNet-50-V1D model outperformed the others. The exception is for observed the standard deviation. We can attribute the superiority of the SCNet-50-V1d model to its robustness due to performing more operations on its self-calibrated convolutions, unlike vanilla convolutions. The SCNet-50-V1D model reached an MSE equal to 4.12 and an MAE equal to 1.24.

The model that comes closest to SCNet-50-V1D was the DenseNet-121 model. It achieved an MSE equal to 7.33 and an MAE equal to 1.67. According to Zhang \textit{et al}. \cite{zhang2021resnet}, DenseNet models are known for their superior feature generalization with fewer parameters than ResNet and other older DL models. ResNet-34 and AlexNet models could not reach error like the SCNet-50-V1D models. While ResNet-34 reached MSE of 12.27 and MAE of 2.31, AlexNet reached MSE of 13.65 and MAE of 2.37. Note that they stand only a little behind in terms of errors from what was reported by the two best models SCNet-50-V1D and DenseNet-121.

Between these two best models and ResNet-34, we find SqueezeNet that achieved an MSE and and an MAE of8.91 and 1.90 respectively. The SqueezeNet model has a smaller amount of parameters compared to ResNet-34, and this fact may have weighed on the results obtained. ResNet-34 models generally converge better when there is a large amount of data used for training \cite{he2016deep}. This may have directly impacted the performance and feature generalization of the model, given that we are facing a learning problem with relatively a small amount of data.

Last but not least, we have the AlexNet model, with an MSE equal to 13.65 and an MAE equal to 2.37, and the VGG-11 model which resulted in am MSE of 27.34 and an MAE value of 3.92. The AlexNet and VGG-11 models contain, in their architectures, several stacked convolutional layers, that is, without concatenation operations or residual operations, as is the case with the DenseNet or ResNet models. The fact of not having transfer features in their layers can impoverish the feature generalization process, causing a more significant error.

\begin{figure}[ht!]
    \centering
    \resizebox{\columnwidth}{!}{\includegraphics{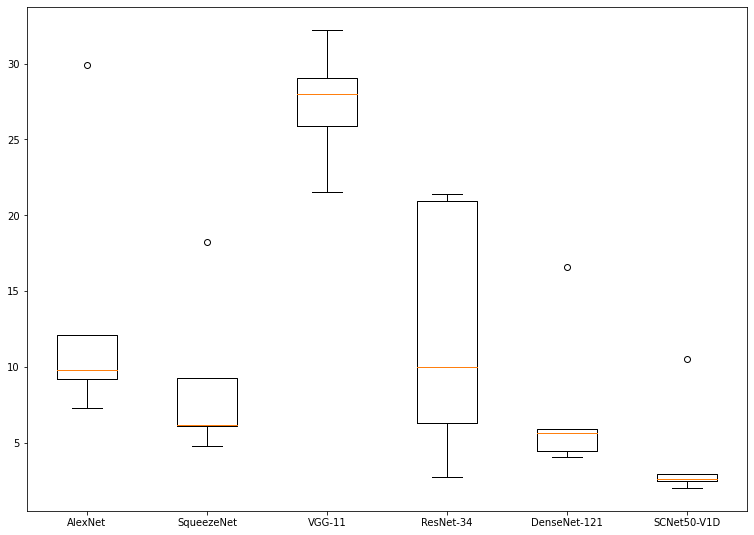}}
    \caption{Boxplot representation for MSE results reached by the DL models.}
    \label{fig:boxplot1}
\end{figure}

All the analysis previously performed considered only the average of the errors of the five cross-validation runs. However, we still need to consider the standard deviation. To carry out a more concise analysis of the behavior of the models, we can also present the results in boxplot form in Fig. \ref{fig:boxplot1}. We consider the MSE error reached by each model into the boxplot.

One can observe the error variance of each of the analyzed models. VGG and ResNet models reached the highest standard deviation and also provided a more significant variation in the boxplot. The other models did not show much variation, except for the presentation of outliers, which impacted the increase in the standard deviation for the models in question.

Regarding outliers, we can observe this behavior in the AlexNet, SqueezeNet, ResNet-34, and SCNet-50V1D models. This behavior may have happened coincidentally in one of the cross-validation executions, which impacts that the data used in training causing poor learning in the models compared to the other executions. The VGG-11 and ResNet-34 models showed, in general, poor learning performances in all executions. For this reason, they may not present outliers in their executions.

The boxplot analysis gives us even greater certainty about the models that performed better in learning. We can see that the DenseNet-121 and SCNet-50-V1D models do not overlap with other models in the boxplot. More specifically, we can see that the SCNet-50-V1D model does not overlap with the DenseNet-121 model, which shows that the samples are statistically different. What could counter this argument are the outliers that both models have in common. However, the error obtained in the outlier with SCNet-50-V1D was lower.

This last statistical analysis reinforces the argument that the SCNet-50-V1D model provides better results for detecting the pose of the robotic arm. Furthermore, we can verify that SCConvs offer better results for a regression task, in addition to the functions indicated by the model proponents, Liu et al. \cite{liu2020improving}. In this work, SCConvs contributed to the construction of feature representation at different scales through self-calibrating operations, contributing to a better generalization of the data. It was known in the literature that SCConvs could present better results in large-scale problems with a high amount of data. According to the experiments presented in this work, there is evidence that SCConvs can also provide better representation and generalization of data in small-scale problems.

\begin{figure}[ht!]
    \centering
    \resizebox{\columnwidth}{!}{\includegraphics{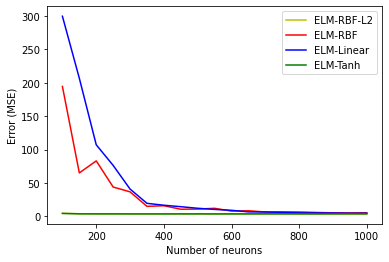}}
    \caption{Relation between the number of neurons and the error reached by the ELM with different activation.}
    \label{fig:elmgraph}
\end{figure}

Previously, we presented the results for detecting the robotic arm pose using only the DL models. Next, we analyze the impact of using an ELM network to refine the results obtained by the SCNet-50-V1D model. Fig. \ref{fig:elmgraph} presents a graph containing the number of neurons \textit{vs.} error (MSE) reached by the specific ELM network.

The refinement of the poses of the robotic arm with ELM can occur with any activation function for the neurons. However, the number of neurons is an essential factor that must be considered for the refinement step. The ELM networks with RBF-L2 kernel and Tanh as activation functions presented similar MSE results during all the variations of the number of neurons. While, ELM networks with Linear kernel and RBF provide results closer to the other two mentioned above, using more than 700 neurons.

By using 1000 neurons, we observe that the results of all ELM networks are similar. However, it is worth mentioning that, if the production environment chooses to use the more lightweight ELM model, the correct option would be to use ELM with RBF-L2 kernel or Tanh with a smaller number of neurons. These last two activation functions provide similar results when using 1000 neurons with the other activation functions. On the other hand, if one chooses to use the ELM network with better learning performance, a more detailed analysis of the obtained results is necessary.

For a better and fairer analysis, we equaled the number of neurons to 1000 for the use of all activation functions. In this analysis, we consider the MSE and MAE metrics. Table \ref{tab:resultsB} presents the results obtained with this analysis, also considering 5-fold cross-validation.

\begin{table}[ht!]
\centering
\caption{MSE and MAE results reported by each activation functions in the ELM network for robotic arm pose refinement.}
\label{tab:resultsB}
\begin{tabular}{lll}
\hline
Model                    & MSE       & MAE       \\ \hline
SCNet50-V1D              & 4.12±3.20 & 1.24±0.13 \\ \hline
SCNet50-V1D + ELM-Linear & 3.76±3.11 & 1.15±0.14 \\
SCNet50-V1D + ELM-RBF    & 5.24±3.47 & 1.47±0.13 \\
SCNet50-V1D + ELM-RBF-L2 & 3.74±3.20 & 1.14±0.13 \\
SCNet50-V1D + ELM-Tanh   & 4.73±3.19 & 1.37±0.13 \\ \hline
\end{tabular}
\end{table}

We observed that ELM with RBF-L2 kernel and Linear reached the smallest errors, considering only an average of the errors obtained in each validation fold. The ELM with the RBF-L2 kernel achieved an MSE of 3.74 and an MAE of 1.14, while the ELM with Linear activation achieved an MSE of 3.76 and an MAE of 1.15. Both generate results that are very close to each other and below those for SCNet-50-V1D (Table \ref{tab:resultsA}). RBF-L2 provides a squared error-based regularization for regression problems, which may explain its smaller errors.

We also analyze ELM with RBF kernel, which achieved an MSE of 5.24, and an MAE of 1.47, while with Tanh it reached an MSE of 4.73 and an MAE of 1.37. Observe that both configurations worsened the error results that SCNet-50-V1D achieved. Despite always obtaining error results that varied little regardless of the number of neurons (compared to ELM-RBF and ELM-Linear), ELM with Tanh activation did not show interesting results with 1000 neurons. The distribution of neurons into 1000 data may have impacted the low results of Tanh (which is more suitable for binary classification) and RBF (that could not provide a better approximation of the function).

\begin{figure}[ht!]
    \centering
    \resizebox{\columnwidth}{!}{\includegraphics{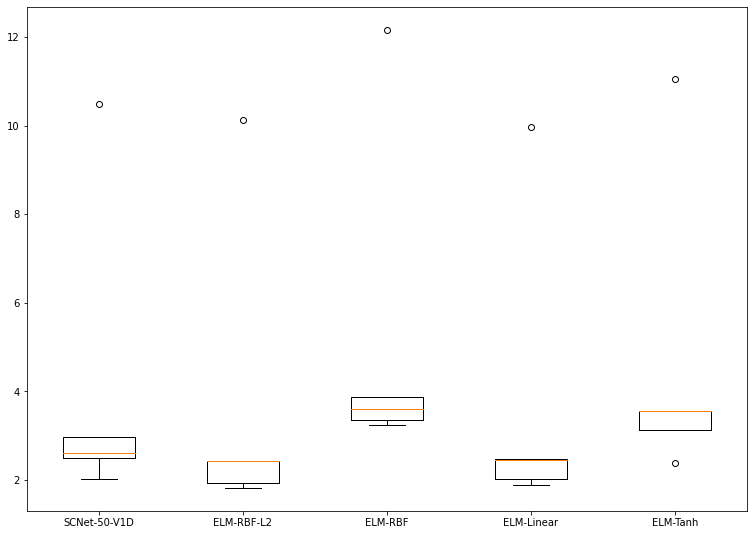}}
    \caption{Boxplot representation for MSE results reached by SCNet-50-V1D with and without ELM network.}
    \label{fig:boxplot2}
\end{figure}

Considering an analysis from the standard deviation viewpoint, we can see that ELM with RBF-L2 kernel and Linear do overlap. Fig. \ref{fig:boxplot2} shows a boxplot with all the results obtained by SCNet-50-V1D with ELM. For comparison purposes, the boxplot also shows the results of SCNet-50-V1D without ELM.

We verified that, regarding the models with ELM, the best learning performance with RBF-L2 and Linear is confirmed. These two models obtained smaller error results in all five execution runs, while not denoting any overlap with the ELM models with RBF and Tanh. In addition, the outlier that had been previously evidenced only with SCNet-50-V1D appears again, proving that in one of the execution rounds, there was indeed some difficulty in the modeling and learning process of the models. In addition, the ELM with Tanh, in particular, presented a second outlier, however with a smaller MSE, while having a significant convergence in one of the validation folds.

The boxplot analysis also indicates that it is impossible to state which of the models is better, ELM with RBF-L2, or Linear ELM. This is because result's samples are not statistically different. Both models reached similar results and behavior across all the validation fold, as reflected by the MSE. Still, about these last two models, their results generally do not overlap with the results of the SCNet-50-V1D model without ELM. This phenomenon demonstrates that the samples may not be statistically different and that the use of ELM networks refines the robotic arm poses regression error results. Improving detection results is essential for a better representation of poses, as wrongly detected keypoints can negatively impact the results, causing risks in a critical HRC system.

\subsection{Robotic arm movement prediction}

In this section, we present the experimental results for predicting the future movement of the robotic arm. First, we start by showing the results using the LSTM model. Then we present the results related to the GRU model. Due to the many results for each model, we opted to show only MSE results. All experiments considered SCNet-50-V1D with ELM-RBF-L2 as a robotic arm pose detector over the video recording.  

Table \ref{tab:resultsD} shows the results obtained for predicting the robot's future movement using only the LSTM model for each configuration defined in the grid search experiments. 

\begin{table}[ht!]
\centering
\caption{MSE results reported by LSTM model for robotic arm movement prediction.}
\label{tab:resultsD}
\begin{adjustbox}{max width=\linewidth}
\begin{tabular}{
>{\columncolor[HTML]{FFFFFF}}l 
>{\columncolor[HTML]{FFFFFF}}l 
>{\columncolor[HTML]{FFFFFF}}l 
>{\columncolor[HTML]{FFFFFF}}l 
>{\columncolor[HTML]{FFFFFF}}l 
>{\columncolor[HTML]{FFFFFF}}l 
>{\columncolor[HTML]{FFFFFF}}l 
>{\columncolor[HTML]{FFFFFF}}l }
\hline
\cellcolor[HTML]{FFFFFF}                              & \multicolumn{7}{l}{\cellcolor[HTML]{FFFFFF}Future Window} \\ \cline{2-8} 
\multirow{-2}{*}{\cellcolor[HTML]{FFFFFF}Past window} & 1      & 5      & 15     & 30    & 60    & 90    & 120    \\ \hline
10                                                    & 28.89  & 27.14  & 33.27  & 40.81 & 59.46 & 74.02 & 89.69  \\
20                                                    & 32.61  & 29.95  & 36.31  & 40.03 & 60.98 & 77.65 & 100.03 \\
30                                                    & 25.69  & 32.79  & 31.86  & 44.02 & 51.01 & 71.69 & 93.87  \\
45                                                    & 34.19  & 33.38  & 33.11  & 39.51 & 61.49 & 72.38 & 101.39 \\
60                                                    & 29.65  & 27.56  & 37.24  & 42.62 & 49.88 & 71.88 & 98.97  \\ \hline
\end{tabular}
\end{adjustbox}
\end{table}

The results show that the future window parameter has a significant impact on the results. We first highlight the use the  prediction window values of 1 and 5, because in some cases, prediction with a window equal to 5 outperformed prediction using a window equal to 1, independently of almost all sizes of the past data window, except when this is equal to 30. This demonstrates one of the characteristics of LSTM: it is easier to learn long-term data (despite the the use of a small dataset) \cite{hochreiter1997long}. We also highlight that the LSTM algorithm presents good results of future prediction of the robot with a small past window with a future window equal to 5. The smaller the past window, the less data will be processed and propagated in the network, which may decrease the amount of processing required to arrive at a low MSE result.

Despite highlighting the prediction window of 5, LSTM tended to increase MSE as the prediction window size is more significant, regardless of the size of the past data window. MSE results start to increase considerably after using 30 frames (or 1 second) in the future forecast window. Despite this, we can still observe that the use of a past window equal to 10 provided better forecast results, in general, except against the past window similar to 30 (in some cases). Nonetheless, the results obtained in these two prediction windows remained very close.

Table \ref{tab:resultsC} shows the results obtained for predicting the robot's future movement using only the GRU model for each configuration defined in the grid search experiments.

\begin{table}[h]
\centering
\caption{MSE results reported by GRU model for robotic arm movement prediction.}
\label{tab:resultsC}
\begin{adjustbox}{max width=\linewidth}
\begin{tabular}{
>{\columncolor[HTML]{FFFFFF}}l 
>{\columncolor[HTML]{FFFFFF}}l 
>{\columncolor[HTML]{FFFFFF}}l 
>{\columncolor[HTML]{FFFFFF}}l 
>{\columncolor[HTML]{FFFFFF}}l 
>{\columncolor[HTML]{FFFFFF}}l 
>{\columncolor[HTML]{FFFFFF}}l 
>{\columncolor[HTML]{FFFFFF}}l }
\hline
                              & \multicolumn{7}{c}{Future Window}  \\ \cline{2-8} 
\multirow{-2}{*}{Past window} & 1 & 5     & 15    & 30    & 60    & 90    & 120   \\ \hline
10                                                    & 15.58      & 20.36 & 21.25 & 26.27 & 40.72 & 57.43 & 82.49 \\
20                                                    & 20.54      & 21.91 & 21.67 & 24.50 & 42.89 & 59.90 & 83.84 \\
30                                                    & 18.07      & 19.76 & 17.85 & 26.59 & 43.63 & 58.53 & 85.25 \\
45                                                    & 16.49      & 18.64 & 22.20 & 26.26 & 37.58 & 62.38 & 81.75 \\
60                                                    & 18.77      & 16.58 & 20.29 & 24.31 & 38.38 & 65.47 & 83.05 \\ \hline
\end{tabular}
\end{adjustbox}
\end{table}

First, we can highlight the superiority of GRU over LSTM in our experiments. A likely explanation for this phenomenon is that the fact that the small database may have weighed in favor of the GRU \cite{yang2020lstm}. The GRU model is more straightforward than LSTM, having only two gates. It is more likely to obtain better results when using less data. On the other hand, LSTM tends to provide a better outcome when exposed to a large amount of data (total number of samples in the dataset). As observed with the LSTM model, the GRU also presented better MSE results with prediction windows equal to 1 and 5 regardless of the past data window, except when this is equal to 30.

Similar to the LSTM, the GRU model tends to increase the error as we increased the size of the prediction window of the future. With the increasing of the future prediction window, the error can also to increase \cite{patel2020deep}. The error tendency in the first frames of the prediction sequence is smaller, and as you move away from the starting point of the prediction window, the error increases.

Unlike the pose detection demonstrated in the previous subsection, future prediction does not consider images as input. The only robot poses data in a grouping of frames from the past, making the inference in a “blindly way”. In addition, the MSE metric penalizes the highest error results (farthest frames) during inference. This fact also explains why the MSE results were much higher than the robot poses detection.

With the results obtained by the GRU, we can also highlight the competitiveness of the results when we use a past window equal to 10, obtaining results close to those of other past data windows. In addition, it is also highlighted that the GRU promoted a significant increase in MSE when prediction windows equal to or greater than 60 were used. This result can be considered another advantage for LSTM.

%% file: sections/conclusion.tex
\section{Conclusion}
\label{sec:conclusion}

In this paper, we proposed a new framework for robotic arm pose estimation and future movement prediction in the context of human-robot collaboration in a well-controlled scenario. This framework consists of two modules, the first module estimates robotic arm pose using self-calibrated convolutions and ELM, while the second module predicts its future movement.

Results listed in this paper suggest that our framework provides satisfactory results with a low detection error. It was possible to reach an MSE of 4.12 using the SCNet-50-V1D, while the proposed model SCNet-50-V1D+ELM reached an MSE of 3.74. It outperforms all analyzed baselines. In this paper, we showed that the use of ELM with self-calibrated convolutions can provide low error results or better generalization for the regression task. Also, the results reached by the LSTM and GRU could ensure low errors for robotic arm movement prediction.

Although this work contributes to risk mitigation is a HRC, it does not claim solving the  collision risk problem in its entirety. Our goal was to address the open gaps previously presented and propose a new framework that mainly provides future movement prediction and detect the pose of the robotic arm. This last mentioned step can be essential for risk assessment in the HRC context. The use of this framework allows for example a human agent to know where a robot will move to, providing it ample time for making decisions and reacting in order to remain safe and avoid any damage to a moving robot and humans nearby.

As future work, we intend to extend our approach by using the proposed models for risk evaluation in a well-controlled scenario. We plan to use the models for a joint human and robotic pose estimation, besides using deep learning forecasting models to analyze the extracted keypoints to predict possible risk or collision situations in the human-robot collaboration. These extensions stand to enable us to map future collision situations even before they occur. We also intend to extend our mechanism to work in a more advanced and complete environment with more people, robots, and interactions in order to develop a more robust and realistic system for collision detection.